\documentclass[conference]{IEEEtran}

\usepackage[cmex10]{amsmath}

\usepackage{color}
\usepackage{ulem}

\usepackage{epsfig,multirow}
\usepackage{ifpdf}

\newif\ifpdf
\ifx\pdfoutput\undefined
   \pdffalse
\else
   \pdfoutput=1
   \pdftrue
\fi
\ifpdf
   \usepackage{graphicx}
   \usepackage{epstopdf}
\else
   \usepackage{graphicx}
\fi

\usepackage{amssymb}

\usepackage{algorithm2e}

\usepackage{stmaryrd}

\begin{document}

\title{Storing sequences in binary tournament-based neural networks}

\author{\IEEEauthorblockN{Xiaoran Jiang, \textit{Member, IEEE},
    Vincent Gripon, \textit{Member, IEEE}, Claude Berrou, \textit{Fellow, IEEE,} and \\ Michael Rabbat, \textit{Member, IEEE}}
}

\maketitle

\begin{abstract}
%\boldmath
An extension to a recently introduced architecture of clique-based neural
networks is presented. This extension makes it possible to store sequences with high efficiency. To obtain this property, network connections are provided with
orientation and with flexible redundancy carried by both spatial and temporal redundancy, a mechanism of anticipation being introduced in the model. 
In addition to the sequence storage with high efficiency, this new scheme also offers biological plausibility.
In order to achieve accurate sequence retrieval, a double layered structure combining
hetero-association and auto-association is also proposed.\\
\end{abstract}

\begin{keywords}
Associative memory, sequential memory, sparse coding, information theory, redundancy, directed graph
\end{keywords}

\let\thefootnote\relax\footnote{This work was supported by the European Research Council under Grant
ERC-AdG2011 290901 NEUCOD. The first three authors are with the Electronics Department, T\'el\'ecom Bretagne, Brest 29238, France, and also with the Laboratory for Science and Technologies of Information, Communication and Knowledge, Brest 29238, France (e-mail: xiaoran.jiang@telecom-bretagne.eu; vincent.gripon@telecom-bretagne.eu; claude.berrou@telecom-bretagne.eu). Michael Rabbat is with the Department of Electrical and Computer Engineering, McGill university, Montreal, Canada (e-mail: michael.rabbat@mcgill.ca). The contribution of Michael Rabbat in this study was carried out while he was visiting T\'el\'ecom Bretagne.}

\section{Introduction}
Learning and storing temporal sequences in neural networks have long been an important research topic since the forward progression of time is a fundamental aspect of human reasoning. Different approaches have already been proposed. Some exploit the dynamics of spiking neurons  [\ref{Barber}][\ref{Brea}], others are based on binary neurons, similar to the McCulloch-Pitts model [\ref{McPitts}] with binary inputs and outputs. 

Most learning theoretical models are based on the Hebbian assumption, that is, changes in synaptic strength related to correlations of pre- and post-synaptic firing. Among them, Hopfield networks [\ref{Hopfield}], known predominantly for their ability to encode static patterns, could be adapted to store temporal sequences in an incremental manner. For example, in [\ref{Maurer}], the Hopfield model is extended to encode a time series of patterns. The sequences are stored in a set of Hopfield networks linked to one another through a matrix of weights. However, the limitation of Hopfield networks is well known: only ``about $0.15N$ states can be simultaneously remembered before error in recall is severe'', $N$ being the number of units in the network [\ref{Hopfield}]. This is due to ``catastrophic interference" (CI) [\ref{Mccloskey}] or ``catastrophic forgetting" (CF) [\ref{French}]. Indeed, in Hopfield-like networks, the storage of new information affects on all the connections and thus introduces noise to all the messages that have been stored. Other models such as those in [\ref{Hinton}] and [\ref{Sutskever}] are based on Boltzmann machines [\ref{Ackley}], in which the weights are continuous, and these models have a similar issue with new messages affecting the ability to retrieve older ones.

In contrast to the models discussed above, Willshaw-type models [\ref{Willshaw}][\ref{Sommer}] consider sparse messages and binary connections instead of weighted ones. The CI issue is then better resolved: a newly stored pattern only involves a small number of connections compared to the network size. It can partially overlap with some older ones, but the number of overlapped patterns is largely smaller than the total number of ones that the network can potentially store. Concerning sequential learning, it is possible to store the most elementary sequences (those containing only two patterns, a cue pattern and a target pattern) in a bipartite Willshaw network. A sparse pattern \textit{a} represented in memory \textit{x} is associated with another sparse pattern \textit{b} in memory \textit{y} via hetero-associative connections. A one-step retrieval algorithm enables to retrieve \textit{b} provided \textit{a}.

By introducing an organization in clusters, a recently proposed Willshaw-type neural network based on cliques and sparse representations [\ref{GB1}] can treat information with non binary alphabets. This is extended in [\ref{Aliabadi}] with a further sparse organization and structure. This model demonstrates large storage diversity (the number of storable messages with a relatively high recovery rate) and capacity (the amount of storable information), as well as strong robustness with respect to erasures and errors. However, the non-oriented connections in this model are not biologically relevant. It is then of high interest to replace them with oriented ones. Orientation of connections would naturally offer the network the ability to store sequential information. One may expect this structure to inherit the good properties of the clique-based networks to solve some well-known issues such as CI in sequence learning.

Anticipation is important in human sequential behavior. For instance, it is shown in [\ref{Synder}] that for sound sequences, induced gamma-band activity predicts tone onsets and even persists when expected tones are omitted. Subsequently, [\ref{Leaver}] suggests that learning sound sequences recruits similar brain structures as motor sequence learning, in which retrieving stored sequences of any kind involves predictive readout of upcoming information before the actual sensorimotor event. Anticipation in visual graphemic sequences is studied in [\ref{Orliaguet}], in particular the between-letter context effects. It is shown that this kind of multi-model approach such as \textit{shape + movement} anticipates better than a simple \textit{shape} or \textit{movement}. Inspired from these experiments, the sequential anticipation in this paper exploits redundancy in two dimensions: temporal and spatial redundancy. 

The rest of paper is organized as follows: Section \ref{Chpt_clique_based_network} recalls the principles of storing fixed length messages in clique-based networks, which is at the root of the work presented in this paper. In Section \ref{Chpt_chain_of_tournmt}, the oriented sparse neural networks based on original oriented graphs, called ``chains of tournaments" are demonstrated to be good material to store sequential information with an anticipated mechanism. Generalization is proposed in Section \ref{Chpt_learning_seq_pat}. Section \ref{dbl_layer_strct} introduces a structure combining clique-based and tournament-based networks in order to obtain accurate retrieving. Finally, a conclusion is provided in Section \ref{conclu}.

\section{Clique-based neural networks}
\label{Chpt_clique_based_network}
\subsection{Summary}
Let us recall the key points of the clique-based neural networks initially introduced in [\ref{GB1}] and then extended in [\ref{Aliabadi}] which could be seen as a Willshaw structure [\ref{Willshaw}][\ref{Sommer}] with the ability to store messages with a non binary alphabet. Such a network is composed of $n$ binary nodes organized in $\chi$ clusters.  Each cluster contains a certain number of nodes, which materialize the alphabet. Different clusters may reify different alphabets. For example, a cluster of 26 nodes may represent the 26 letters of the English alphabet and another cluster of 100 nodes, in the same network, may represent the decimal numbers from 1 to 100. For the reason of simplicity, unless specifically mentioned, in the sequel we consider clusters of same size $l=n/\chi$ each. We denote by $n_{ij}$ the $j^{\text{th}}$ node in the $i^{\text{th}}$ cluster, the value of which $v(n_{ij})$ is respectively one or zero, activated or not. In our vocabulary, the
nodes are called ``fanals'', since during the storage process, only one of them in the same cluster can be activated at the same time.

\begin{figure}[!t]
\centering
\includegraphics[width=3.5in]{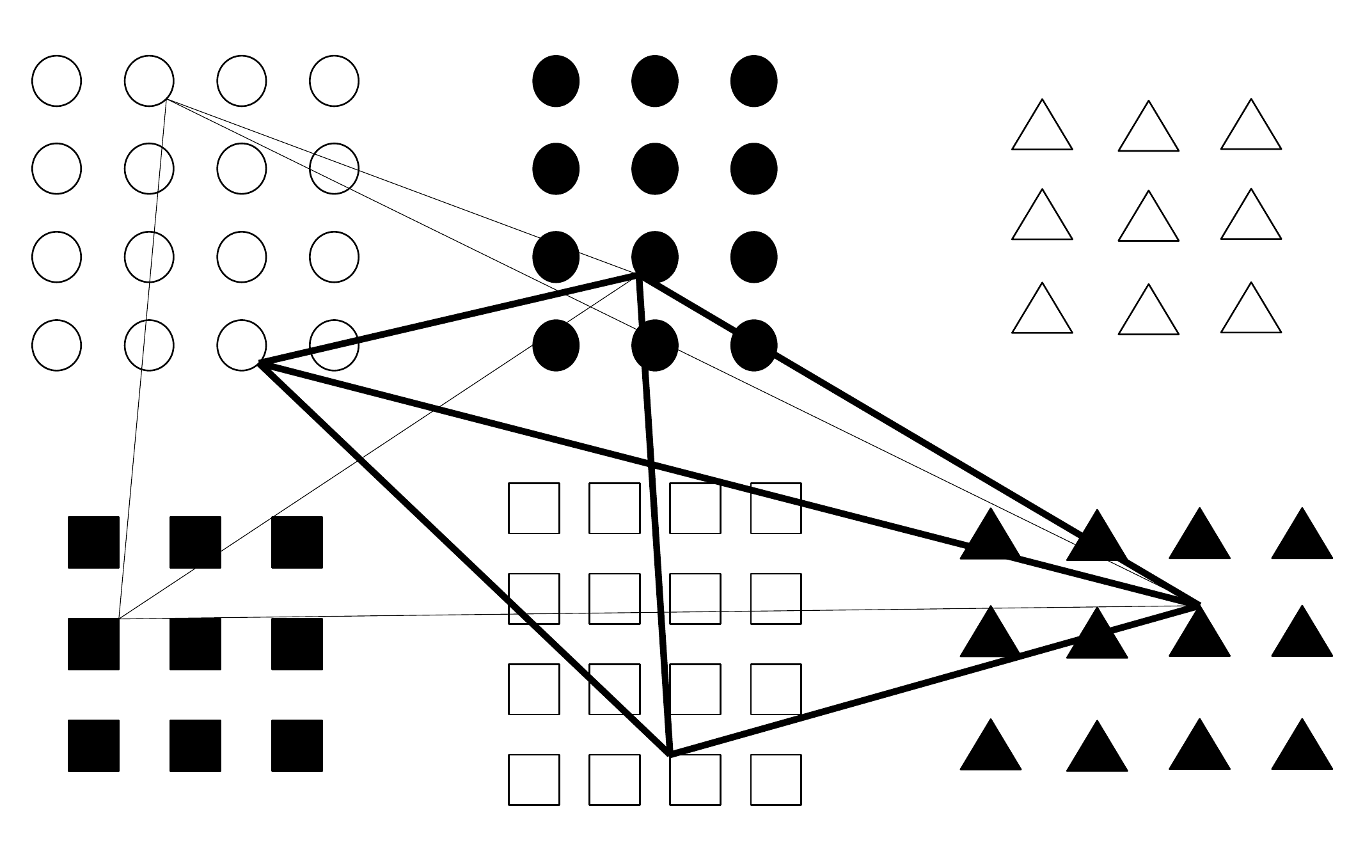}
\caption{Storing process illustration for non-oriented clique-based networks. Two messages of order $c=4$ are stored in a network of $\chi=6$ clusters in form of cliques. One edge is shared. Fanals are represented respectively in 6 clusters by open circles, filled circles, open triangles, filled rectangles, open rectangles and filled triangles.}
\label{Non_oriented_network}
\end{figure}

Let $\mathcal{M}$ be the set of messages to be stored by the network. Each message $m \in \mathcal{M}$ is split into $c$ sub-messages or symbols $m^{i}$. Graphically, each sub-message is mapped to a unique fanal in the corresponding cluster.  As a consequence, storing a message is equivalent to storing the corresponding pattern of $c$ fanals. This pattern is then represented by a clique, that is, a set of fanals such that each one is connected to the others.  It has been proven in [\ref{GB2}] that cliques are codewords of a good error correcting code. The binary edge weight between nodes $n_{ij}$ and $n_{i'j'}$ is denoted by $w_{(ij)(i'j')}$. The weight of an existing connection remains unchanged even if the same pair of fanals appears in two or more messages. An example with two messages of order $c=4$ stored in a network of $\chi=6$ clusters is represented in Fig. \ref{Non_oriented_network}. 

Formally, if we denote $W(m_{k})$ the connection set of the corresponding clique after learning message $m_{k}$, the connection set of the associated graph after learning $\{m_{1}, m_{2}, ..., m_{M}\}$ can therefore be defined by the union:
\begin{equation}
\label{learning_fixed_length}
W\{m_{1}, m_{2}, ..., m_{M}\} = \bigcup_{k=1}^{M} W(m_{k}).
\end {equation}

At the retrieval stage, the network is provided with a message $\hat{m}$, possibly a distorted version of the stored message $m$. Several forms of distortions are envisaged. We call an \textit{erasure}, the situation when a cluster that should be active is not provided with information, in other words none of the fanals in this cluster are activated;  an \textit{error}, the situation when the fanal activated in a cluster is not the right one; and an \textit{insertion}, the situation when some fanals are activated in the clusters that should be silent. Based on the set of existing connections, an appropriate fault tolerant decoding algorithm should be able to give an estimate $\tilde{m}$, which would be the nearest from $m$ in terms of Hamming distance among $\mathcal{M}$: minimum Hamming distance amounts maximun correspondence between two graphical patterns. 

The decoding procedure could be iterative and each iteration normally consists of two steps:  \textit{message passing} and \textit{selection of the winner} [\ref{Ala}]. The former exchanges stimuli (that is, some of the sub-messages $m^i$) within the whole network via established connections, and the contributions are added at each fanal. The latter chooses the fanals that are the most likely correct for the next iteration. In the case that all the clusters are addressed by a single message ($\chi = c$), a winner-take-all (WTA) rule is performed locally within each cluster [\ref{GB1}].  This rule is adapted in [\ref{Aliabadi}] for storing sparse messages ($\chi  \gg c $) by performing Global winner-take-all (GWTA) in the whole network. 

However, GWTA is not suitable for an iterative decoding process. In fact, as mentioned in the original paper [\ref{Aliabadi}], although there is indeed an advantage to so called iterative "guided decoding" (the indices of the clusters involved in the target message are known beforehand, which reduces to the similar situation as in [\ref{GB1}]), on the contrary, iterations do not help for ``blind decoding'' (when the cluster indices are not known beforehand). Indeed, GWTA will only choose the fanals strictly with the highest score. Corrupted information at a certain iteration could incorrectly fire very few fanals with the highest score, all the others being turned off. At the next iteration, the very limited information provided to the decoder would activate an excessively large number of fanals. The network would then oscillate between these two states as long as the iterations continue. 

A more detailed study of retrieval algorithms in clique-based networks is carried out in [\ref{Ala}] [\ref{JiangThesis}] where some improvements are proposed. Among them, Global winners-take-all (GWsTA) selects $\alpha$, an expected number of fanals (that is $c$, the clique order, in the ideal case) that have the maximal or near maximal scores. Global Losers-kicked-out (GLsKO) follows an opposite approach. Instead of selecting the fanals with the highest score, it eliminates those with the lowest score. These two algorithms improve considerably the retrieval performance for ``blind decoding'', especially with input distortions of type \textit{error} or \textit{insertion}. The performance approach that of Maximum likelihood (ML) decoding. These algorithms will be discussed and evaluated in Section \ref{glb_decod_schm} and  Section \ref{dbl_layer_strct} with the adaptation to retrieval of sequences of patterns.

\subsection{Biological and information considerations}
The \textit{data representation} of a neural pattern is of critical importance in neural networks. Two opposite approaches are worth mentioning. On one hand, a knowledge element can be materialized by a highly localized element, called the \textit{grandmother cell} [\ref{Bar72}] [\ref{Gro02}] representation. The problem with this theory is the explosion of material if one would represent a very large number of knowledge elements. On the other hand, the principle of \textit{distributed representation} [\ref{RHM86}] states that a specific piece of information is encoded by a unique pattern of activity over a group of neurons. To make it clear, $n$ grandmother neurons can only represent $n$ knowledge elements, whereas this number is $2^n$ using the distributed approach. To our knowledge, there is more and more evidence that memories are stored in the brain using a distributed representation across numerous cortical and subcortical regions [\ref{Eichenbaum}] [\ref{Fuster}], which means that each memory is stored across thousands of synapses and neurons, and each neuron or synapse is potentially involved in thousands of memories. 

Currently, it is generally assumed that neural networks are structured in a columnar organization. At the lowest level, about $100$ neurons form a cortical microcolumn (also called the minicolumn) [\ref{BC02}], which is a vertical cylinder through the six cortical layers of the brain. The diameter of a microcolumn is about $28-40 \mu $m. There are about $2 \times 10^8$ microcolumns in the human neocortex [\ref{JL07}]. Neurons within a microcolumn perform various tasks: short and long range excitation, short range inhibition. From the point of view of informational organization, the microcolumn is likely the most basic and consistent information processing unit in the brain, instead of a single neuron [\ref{Jones}]-[\ref{Cruz}]. As a consequence, we deliberately call a node in our network a ``fanal'' instead of a ``neuron'' to make it clear that a node corresponds to a microcolumn. Then, dozens of cortical microcolumns form a macrocolumn, analogous to the way that nodes compose a cluster. In the literature, the name of this unit is often ambiguous: ``column", ``macrocolumn" and ``hypercolumn" are commonly interchangeable. Finally, several macrocolumns (clusters) are grouped together to contribute to a functional area of the brain. 

The general concept of the brain as an information processing machine underlines the importance of information theory in cognition problems. A number of research efforts [\ref{Miller}]-[\ref{Cowan}] point out that the \textit{chunk} of information, an abstract but nevertheless measurable unit of information, is relevant to several human cognitive features. In the clique-based network, the division of information into clusters is in fact a way for \textit{chunking}. A sparse message may be composed of chunks, the number of which is significantly smaller than the number of clusters in the entire network. And a hierarchical structure could be envisaged to form clusters of clusters as well as chunks of chunks. However, these perspectives are not the concern of this paper.

\subsection{Degenerated cliques and tournaments}
\label{subChpt_degenerated_cliques}
From the information point of view, the memory is robust and durable and therefore must be encoded \textit{redundantly}. The concept of \textit{informational redundancy} was originally defined by Shannon in the context of communication theory [\ref{Shannon}]. Recently numerous studies have proposed to investigate connectivity redundancy of complex brain networks with the introduction of basic principles of graph theory, such as [\ref{Bullmore}] [\ref{Lanzo}]. However, some authors consider that redundancy can limit the amount of information carried by a neural system: ``in neuroscience, redundancy implies inefficiency'' [\ref{Friston}]. In short, redundancy has to be used with moderation and in a subtle way. 

\begin{figure}[!b]
\centering
\includegraphics[width=3.5in]{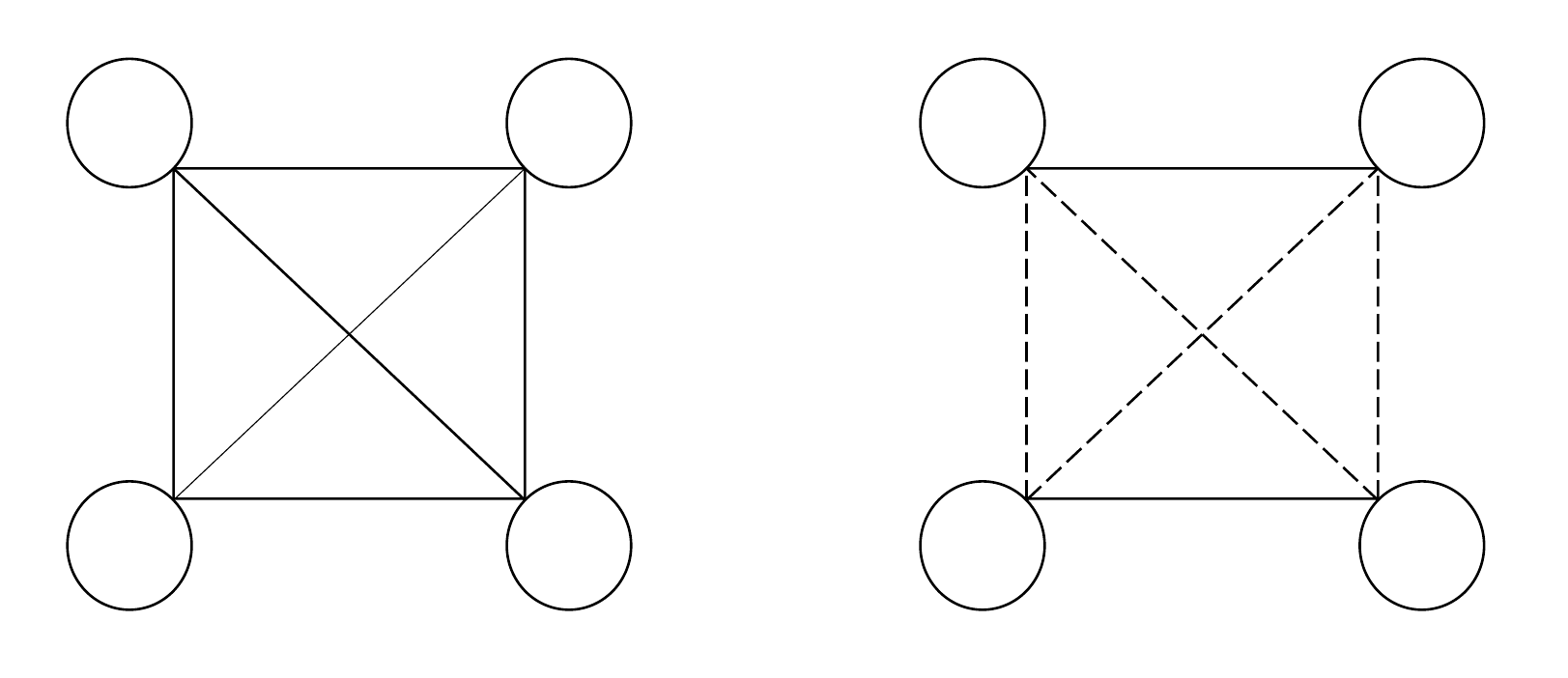}
\caption{A 4-clique has six edges but only two of them are sufficient to specify the pattern.}
\label{clique_redundancy}
\end{figure}

The retrieval algorithms in Section \ref{Chpt_clique_based_network} make use of the high redundancy of 
cliques. As a matter of fact, a clique composed of four nodes for example can be uniquely defined by two connections if well chosen, and the other four
serve as redundant information (see Fig. \ref{clique_redundancy}).

\begin{figure}[!t]
\centering
\includegraphics[width=0.45\textwidth]{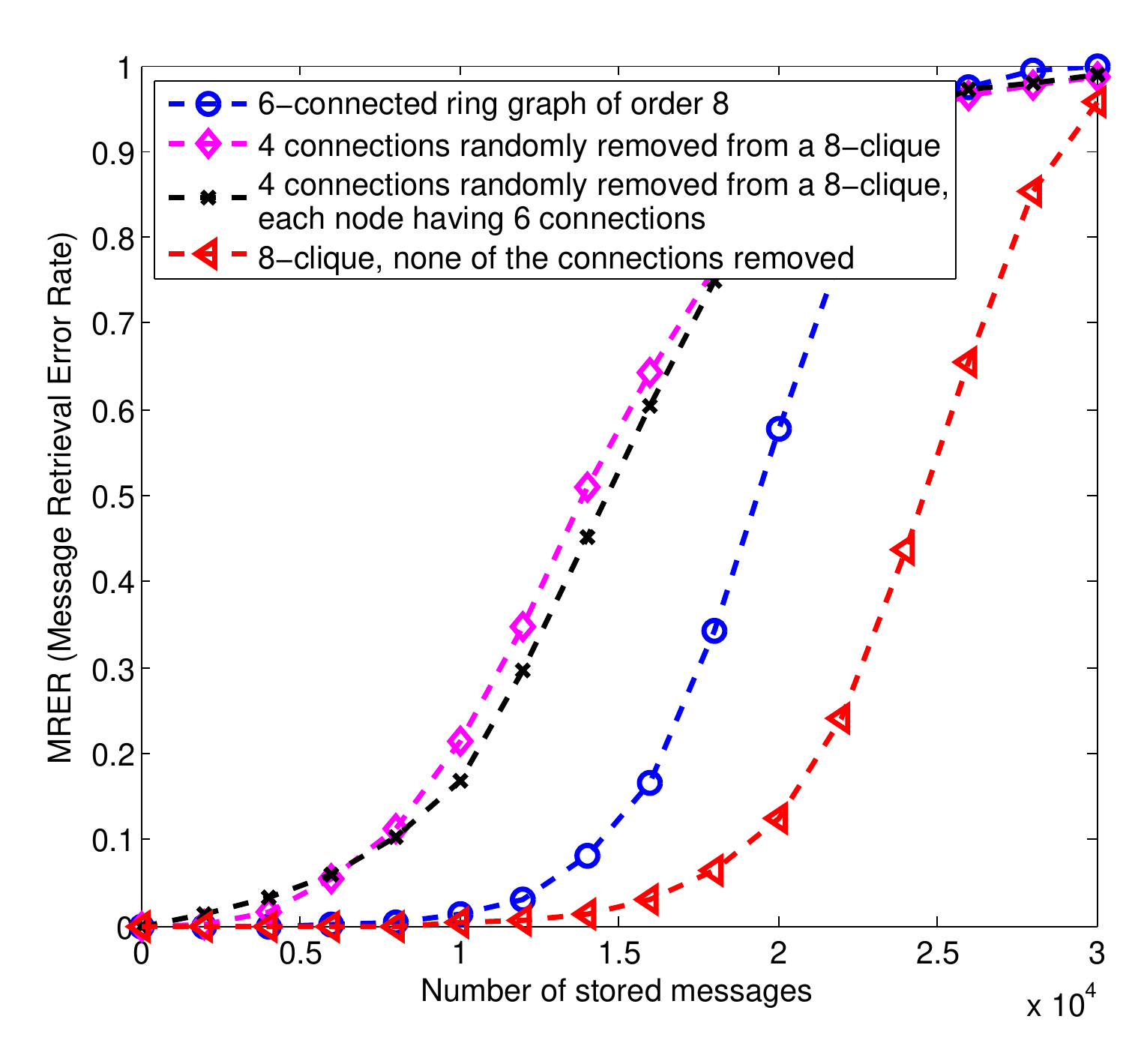}
\caption{Several possibilities to reduce redundancy in a clique. Comparison of the error rate when retrieving a
         partially erased learnt message in function of the
         number of stored messages. The network is composed of 8 clusters of 256 fanals. The information in half of the clusters is not provided and the number of iterations of WTA algorithm is 4.}
\label{simu_clique_redundancy}
\end{figure}

\begin{figure}[!h]
\centering
\includegraphics[width=0.5\textwidth]{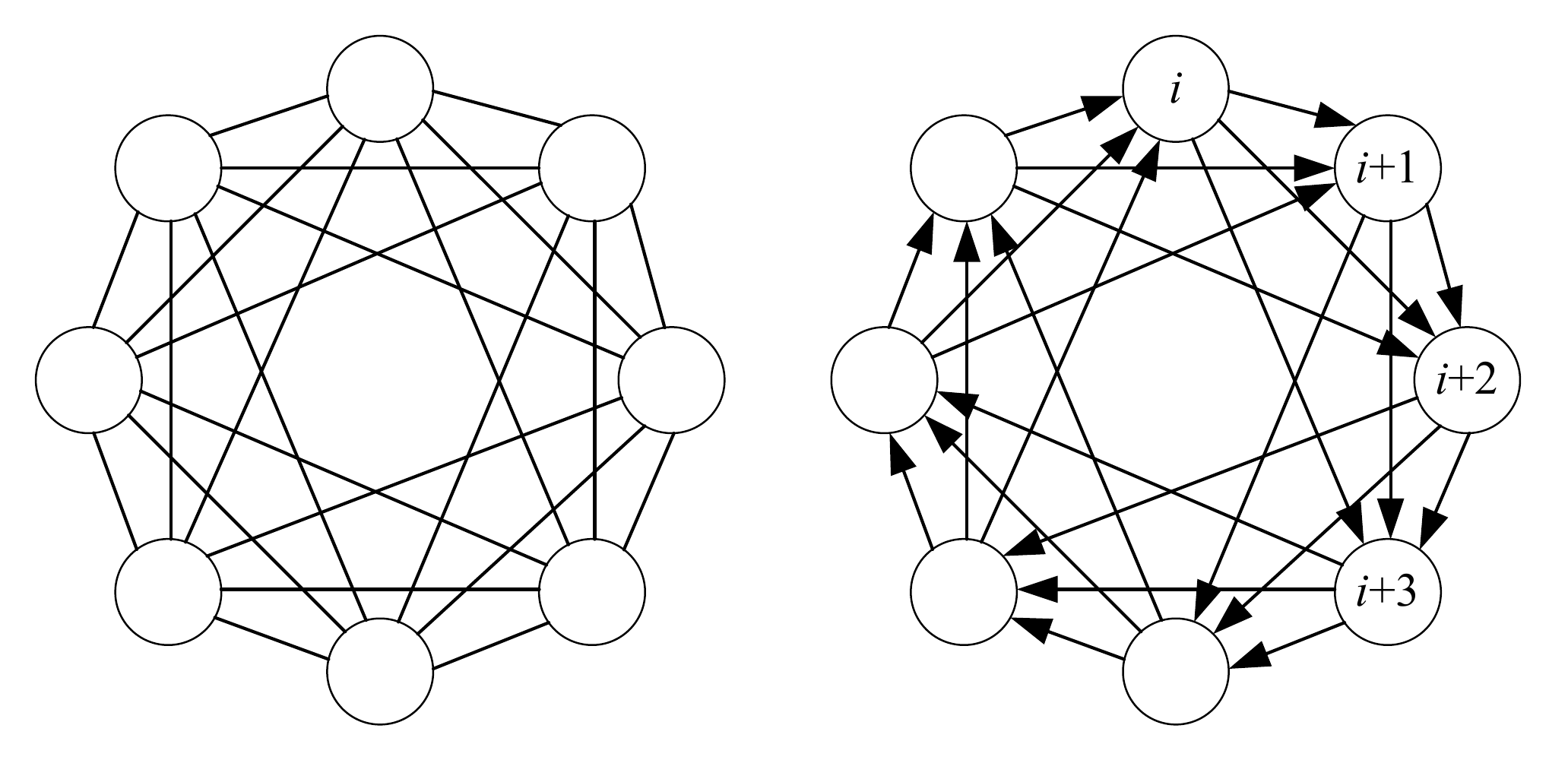}
\caption{A 6-connected ring graph of order 8 (left) and a chain of tournaments of order 8 with incident degree 3. Clusters are represented by circles, and a connection or an arrow represents not a single connection between two fanals, but a set of possible connections between the fanals of two clusters. }
\label{chain_of_cliques_vs_tournaments}
\end{figure}

One may consider degenerating a $c$-clique by taking off some connections. Supposing the total number of connections after this degeneration is a multiple of the order $c$.  We denote this number $rc$, with $r$ an integer and $r<(c-1)/2$.  This degeneration can be achieved in several different ways, going from the inhomogeneous (some nodes having all possible connections whereas some others having very few connections) to the most homogeneous, which is the case for a  $2r$-connected ring graph of order $c$ (see the graph on the left of Fig. \ref{chain_of_cliques_vs_tournaments}), which is denoted by $\mathcal{R}_{r}(c)$ in the sequel. Our simulation (see Fig.\ref{simu_clique_redundancy}) shows that homogeneity of connectivity of the regular degenerated clique, that is,  $\mathcal{R}_{r}(c)$, offers the best retrieval performance.

It has been proven in [\ref{GB1}] that the merit factor of a clique-based code is 2, which means this can be considered a good error correcting code. We can carry out a similar demonstration here for $\mathcal{R}_{r}(c)$. We define the minimum Hamming distance $d_{min}$ of the code based on $\mathcal{R}_{r}(c)$ as the minimum number of edges that differ between two $2r$-connected ring graphs of order $c$.  $d_{min}$ is obtained in the case of only one vertex being different:
\begin{equation}
\label{dmin}
d_{min}=4r.
\end {equation}

When the size of the graph is large in
comparison with that of the pattern, which is here $\mathcal{R}_{r}(c)$, the coding rate $R$ is given by the ratio between the minimum number of edges that are necessary to specify a pattern and the number of edges used. In the case of $\mathcal{R}_{r}(c)$, $R$ can be expressed as:
\begin{equation}
\label{R}
R=\frac{ \lfloor \frac{c+1}{2}  \rfloor}{rc} = \frac{1}{2r} \text{ (for $c$ even).}
\end {equation}

Finally, the merit factor is obtained as:
\begin{equation}
\label{F}
F=R.d_{min}=2.
\end {equation}

We find exactly the same merit factor of 2 as it is for a non degenerated clique-based code: a code based on $\mathcal{R}_{r}(c)$ has a smaller minimal Hamming distance, which is exactly compensated by a higher coding rate.

\section{Tournament-based neural networks}
\label{Chpt_chain_of_tournmt}

\subsection{Structure}
Let us denote $ \llbracket  z_1, z_2 \rrbracket$ the closed integer interval between $z_1$ and $z_2$. In a non-oriented network, the connection matrix is symmetric:
$\forall (i,j),(i',j') \in  \llbracket 1, c  \rrbracket \times \llbracket 1, l \rrbracket,
\omega_{(ij)(i'j')} =\omega_{(i'j')(ij)}$. This is however not
biologically relevant. Indeed, a biological neuron is composed of a cell
body and two types of branched projections: the axon and the
dendrites. The dendrites of a neuron reproduce the electrochemical
stimuli received from the axons of other neural cells to its cell
body, which then propagates the stimuli towards its axon in certain
conditions. This type of propagation of stimuli can only be achieved 
in this unique direction. Considering oriented networks not only provides more biological credibility, but it is also more general, since in a graph, one non-oriented edge is simply the combination of two oriented edges.

Unidirectionality seems, at first glance, to be a handicap to obtain
good performance: it requires twice as much material resources (one has to use two bits in
order to specify the connections between two nodes in an oriented graph,
instead of only one bit in non-oriented one) for lower
redundancy.

If one replaces non-oriented connections by oriented ones in $\mathcal{R}_{r}(c)$, one
obtains a novel structure called ``chain of tournaments" of order $c$ with degree $r$ (c.f. Fig. \ref{chain_of_cliques_vs_tournaments} on the right, $r=3$), which is denoted by $\mathcal{T}_{r}(c)$. In graph
theory, a tournament is a directed graph obtained by assigning a
direction to each edge of a clique. A tournament offers half the redundancy of a clique. And a chain of tournaments is constructed by a succession of a number of tournaments. A chain of tournaments with $r=c-1$ is a $c$-clique.

In the sequel, the parameter $r$ will be called equivalently in several ways: ``incident degree'' as the amount of input information, and ``anticipation degree" as the temporal anticipative span in sequence learning.

\subsection{Storage of sequences of symbols}
\label{storage_seq_sym}
Unidirectionality enables the network to exhibit sequential or
temporal behavior, which opens a novel horizon compared to
non-oriented networks. In fact, the sequential structure imposed by the forward progression of time is omnipresent in all cognitive behaviors. Cognitive sequential learning is unidirectional (irreversible) rather than bidirectional (reversible). Indeed, while learning a song, the succession of the lyrics as well as the melody follows the forward progression of time, and attempting to sing a song in reverse order is extremely difficult if not impossible for human cognitive capacities. In this subsection, we will explain how to store and decode variable length messages whose length is potentially greater than the number of clusters in the network, whereas the length of messages stored in clique-based networks cannot exceed the number of clusters $\chi$.

Beforehand, let us define a temporal sequence following the terminology introduced by Wang and Arbib [\ref{Wang}]. A temporal sequence $s$ of length $L$ is defined as:
\begin{center}
$p_{1} - p_{2} - ... - p_{L} $\\
\end{center}

Each $p_{i} (i=1, 2,..., L)$ is called a component of $s$, which can be just a symbol or a spatial pattern composed of several symbols in parallel. The length of a sequence is the number of components in the sequence. In this section, we are only interested in the sequence of symbols. Sequences of spatial patterns will be discussed in Section \ref{Chpt_learning_seq_pat}.

\begin{figure}[!t]
\centering
\includegraphics[width=0.45\textwidth]{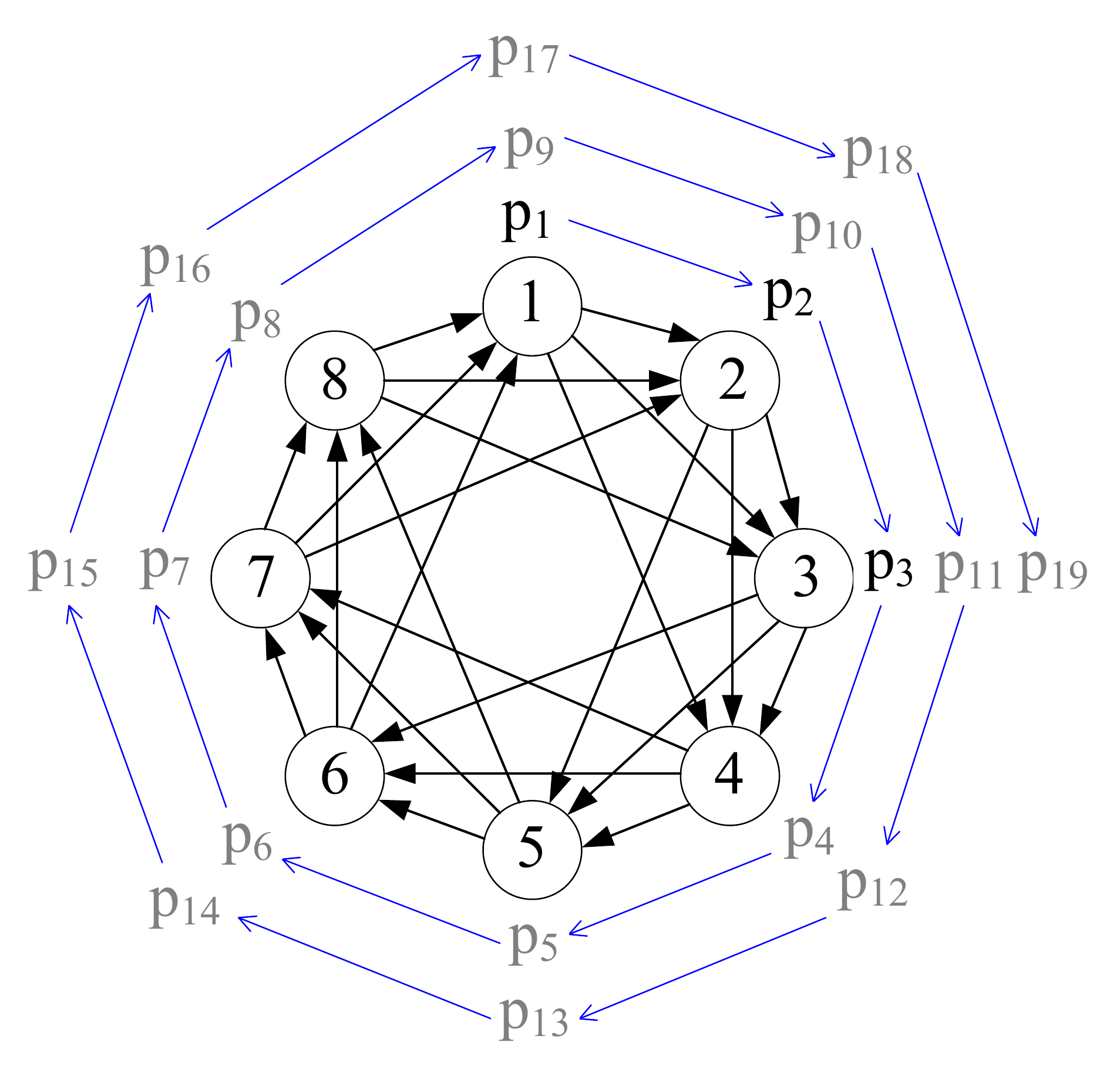}
\caption{A chain of tournaments for storing and decoding arbitrary long sequences. Clusters are represented by circles, and an arrow represents not a single connection between two fanals, but a set of possible connections between two clusters. }
\label{learning_sequence_of_symbols}
\end{figure}

The storage principle is illustrated by Fig. \ref{learning_sequence_of_symbols}. Each symbol $p_{i}$ is mapped to a particular fanal in a corresponding cluster. We define the function $\delta_{i}(i')=(i' - i) \text{ mod } \chi$. We call cluster $i'$ a downstream neighbor of cluster $i$ if $1 \leq \delta_{i}(i') \leq r$. During the storing process, the directed connections are
established successively between the current cluster and its $r$
downstream neighbors. For instance, the fanal corresponding to $p_{1}$ in cluster 1 are connected to those corresponding to $p_{2}, p_{3}$ and $p_{4}$, respectively in clusters 2, 3 and 4; that of $p_{7}$ is connected to those of $p_{8}, p_{9}$ and $p_{10}$, respectively in clusters 8, 1 and 2, etc.

We call that two fanals $n_{ij}$ and $n_{i'j'}$ are \textit{paired} by the sequence $s$ on passage $\pi$, denoted by the binary relation $N_{s, \pi}(n_{ij}, n_{i'j'})=1$, if we have 
\begin{equation}
 \begin{aligned}
	& 1 \leq \delta_{i}(i') \leq r,\\
	& \text{and } \begin {cases}
          f_{i}(p_{i+(\pi-1)\chi}) = n_{ij} \\
          f_{i'}(p_{i'+(\pi-1)\chi}) = n_{i'j'}.
      \end {cases}
\end{aligned}
\end {equation}
Otherwise, $N_{s, \pi}(n_{ij}, n_{i'j'})=0$. Function $f_{i}$ maps the symbol $p_{i+(\pi-1)\chi}$ to the appropriate fanal in cluster $i$ on the $\pi_{\text{th}}$ passage of this sequence. By abuse of language, we admit that words ``symbol" and  ``fanal" are interchangeable in order to simplify notations. 

Thus, after storing a set $\mathcal{S}$ of $S$ sequences of length $L$, the network is defined formally by:\\

$\forall (i,i')  \in \llbracket 1, \chi \rrbracket^{2}, \forall (j,j') \in \llbracket 1, l \rrbracket^{2},$\\
\begin{equation}
w_{(i,j)(i',j')} =
\begin {cases}
1, \text{ if } \exists s \in \mathcal{S}, \exists \pi \in \left \llbracket 1, \frac{L}{\chi} \right \rrbracket, N_{s, \pi}(n_{ij}, n_{i'j'})=1\\
0,  \text{ otherwise}\\
\end {cases}
\end {equation}

Note that the loop structure enables the reuse of the resources. A cluster may be solicited several times, if $L$, the
length of the sequence surpasses the number of clusters $\chi$. In the example illustrated in Fig. \ref{learning_sequence_of_symbols}, cluster 1 is solicited for three times by symbols $p_{1}, p_{9}$ and $p_{17}$. In
this way, the network is able to learn sequences of any length,
not limited by the number of clusters. The encoded sequences can then materialize for example voluminous multimedia streams. 

We suppose that the stored sequences are randomly, uniformly and independently generated among all the possible ones. We assume the independence of connections in order to apply the binomial law. This approximation, verified by simulations, could be intuitively explained by the fact that with some long sequences stored in the network, two connections taken randomly from the graph are unlikely to have been added at the same time. 

The distribution of connections can then be seen as independent Bernoulli variables of parameter $d$, which is the \textit{density} of the network. The value of $d$ can be obtained by binomial arguments. In fact, the probability of having a connection between fanals $n_{ij}$ and $n_{i'j'}$ is by construction the probability of having at least one sequence containing symbol $j$ at position $i$ and symbol $j'$ at position $i'$. For reasons of simplicity, we suppose that $L$ is a multiple of $\chi$. The density of the network after storing $S$ sequences can be expressed as:

\begin{equation}
    \begin{aligned}
   d &=\Pr \left( \exists s \in \mathcal{S}, \exists \pi \in \left \llbracket1,\frac{L}{\chi} \right \rrbracket, N_{s, \pi}(n_{ij}, n_{i'j'})=1\right)\\
                 &=1-\Pr \left(\forall s \in \mathcal{S},\forall \pi \in \left \llbracket 1, \frac{L}{\chi}\right \rrbracket, N_{s, \pi}(n_{ij}, n_{i'j'})=0 \right)\\
                 &=1-\Pr \left(p_{i+(\pi-1)\chi} \neq n_{ij} \lor p_{i'+(\pi-1)\chi} \neq n_{i'j'}\right)^{\frac{SL}{\chi}}\\
                 &=1-\left \{1-\Pr \left(p_{i+(\pi-1)\chi}=n_{ij}\right) \Pr \left(p_{i'+(\pi-1)\chi}=n_{i'j'}\right)\right \}^{\frac{SL}{\chi}}\\
                 &=1 - \left(1 - \frac{1}{l^{2}}\right)^{ \frac{SL}{\chi}}.
 \end{aligned}
\label{den_seq_sym}
\end {equation}

It is important to note that the density is independent of $r$. In fact, the number of potentially established connections for each passage is $r \chi$, which is well dependent on $r$, but the total number of possible connections is also $r$-dependant: $r\chi l^2$. The incremental density for each passage is roughly the ratio between these two, so $r$ is eliminated from the expression. 

\subsection{Error tolerant decoding}
\label{error_tolerant_decoding}
To start the decoding process, the network should be provided with any subsequence of $r$ consecutive symbols, in particular the first $r$ symbols if one would like to retrieve the sequence from the beginning, but this is not necessary. The 3 starting symbols $p_{1}, p_{2}$ and $p_{3}$ are illustrated in black in Fig. \ref{learning_sequence_of_symbols}, the symbols to be deduced being in gray. Note that if the supplied subsequence is in the middle of the sequence, the decoder should be provided with supplementary information where the corresponding clusters are. We see it rather as plausibility than a constraint: for instance, if one gets stuck in the middle of a melody when playing on the piano, it seems easier to recall it by restarting from the beginning of the tune. 

The decoding procedure is sequential and discrete-time such that at each step, the WTA decision is made locally in the following cluster, based on contributions from the $r$ previous clusters. Let us take an example where the trigger sequence is composed of $r$ symbols at the beginning. Formally, the decoding can be expressed as Algorithm \ref{seq_sym_decod}.

\begin{algorithm}[h]
\SetAlgoLined
\SetKwInOut{Input}{input}\SetKwInOut{Output}{output}
\Input{Trigger sequence $s_\text{input}$: $p_{1} - p_{2} - ... - p_{r}$}
\Output{Retrieved sequence $s$: $p_{1} - p_{2} - ... - p_{L}$}
\BlankLine
initialization\;
$s \leftarrow s_\text{input}$\;
\For{$ \lambda \leftarrow 1$ \KwTo $r$}{
	\BlankLine
  	$v(f_{\lambda}(p_{\lambda}) )=1$\;
}
\For{$ \lambda \leftarrow r+1$ \KwTo $L$}{
	\BlankLine
	$i \leftarrow \lambda \text{ mod }\chi + 1$\;
	\BlankLine
         $\forall i, j, v(n_{ij}) \leftarrow  \displaystyle \sum_{1 \leq \delta_{i'}(i) \leq r} \max_{1 \leq j' \leq l}{\left[w_{(i',j')(i,j)}v(n_{i' j'})\right]} $\;
         \BlankLine
         $v_{i}^{\text{max}} \leftarrow \max_{j}\left[v(n_{ij})\right]$\;
         \BlankLine
         $\forall j, v(n_{ij})\leftarrow
        		 \begin {cases}
          		1 \text{ if } v(n_{ij})=v_{i}^{\text{max}}\\
        		  	0 \text{ otherwise}
         		\end {cases}$\;
	\BlankLine
	$p_{\lambda} \leftarrow \{n_{ij}, v(n_{ij})=1\}$\;
	\BlankLine
	$s \leftarrow s - p_{\lambda}$\;
}
\Return{$s$}
\caption{Sequential decoding algorithm in a chain of tournaments: an example when the trigger sequence is composed of $r$ symbols at the beginning.}
\label{seq_sym_decod}
\end{algorithm}

An important comment concerns the anticipation effect included in this decoding principle. Indeed, in Fig. \ref{learning_sequence_of_symbols}, while the decoding is performed in cluster 4, given the results in clusters 1, 2 and 3, activity is already emerging in cluster 5, receiving signals from clusters 2 and 3; and cluster 6, receiving signals from cluster 3. This process of anticipating information gives a rough estimate of the network dynamics in the following time window. This concept resonates with [\ref{Leaver}], according to which the very existence of anticipatory imagery suggests that retrieving stored sequences of any kind involves predictive readout of upcoming information before the actual sensorimotor event.

The clique-based networks are robust towards errors, this property being inherited to some extent by tournament-based networks which are error tolerant in sequential decoding. Indeed, with a sufficient degree of temporal redundancy, specified by $r$, a transitory error would hopefully not propagate to future decoding decisions. However, it is important to note that the benefits of iterative decoding are definitively lost, since only one iteration is applicable in this process.

Let us define the symbol error rate (SBER) as the percentage of the symbols that are not correctly retrieved. One should distinguish two different types of errors: the ones coming from an excessive network density, and thus occuring even if perfect information is provided and which we call \textit{structural} symbol error rate; and those generated because there have been errors within the previous $r$ positions. For an algorithm not tolerant towards errors, as soon as a single error occurs, the next decoding steps will produce errors and as a consequence, practically all errors are of the latter type. One can roughly estimate the structural error rate as the error rate at a single decoding step when perfect information is provided:
\begin{equation}
   P_{e, struct} = 1 - \left(1 - d^{r}\right)^{l - 1}.
\label{innate_sym_err_rate}
\end {equation}

In fact, an error will occur if and only if at least one fanal among $l-1$ in the examined cluster, other than the target one, is also connected to the $r$ previous fanals that are provided correctly. The probability of the existence of such a connection is the network density $d$. 

\begin{figure}[!t]
\centering
\includegraphics[width=0.45\textwidth]{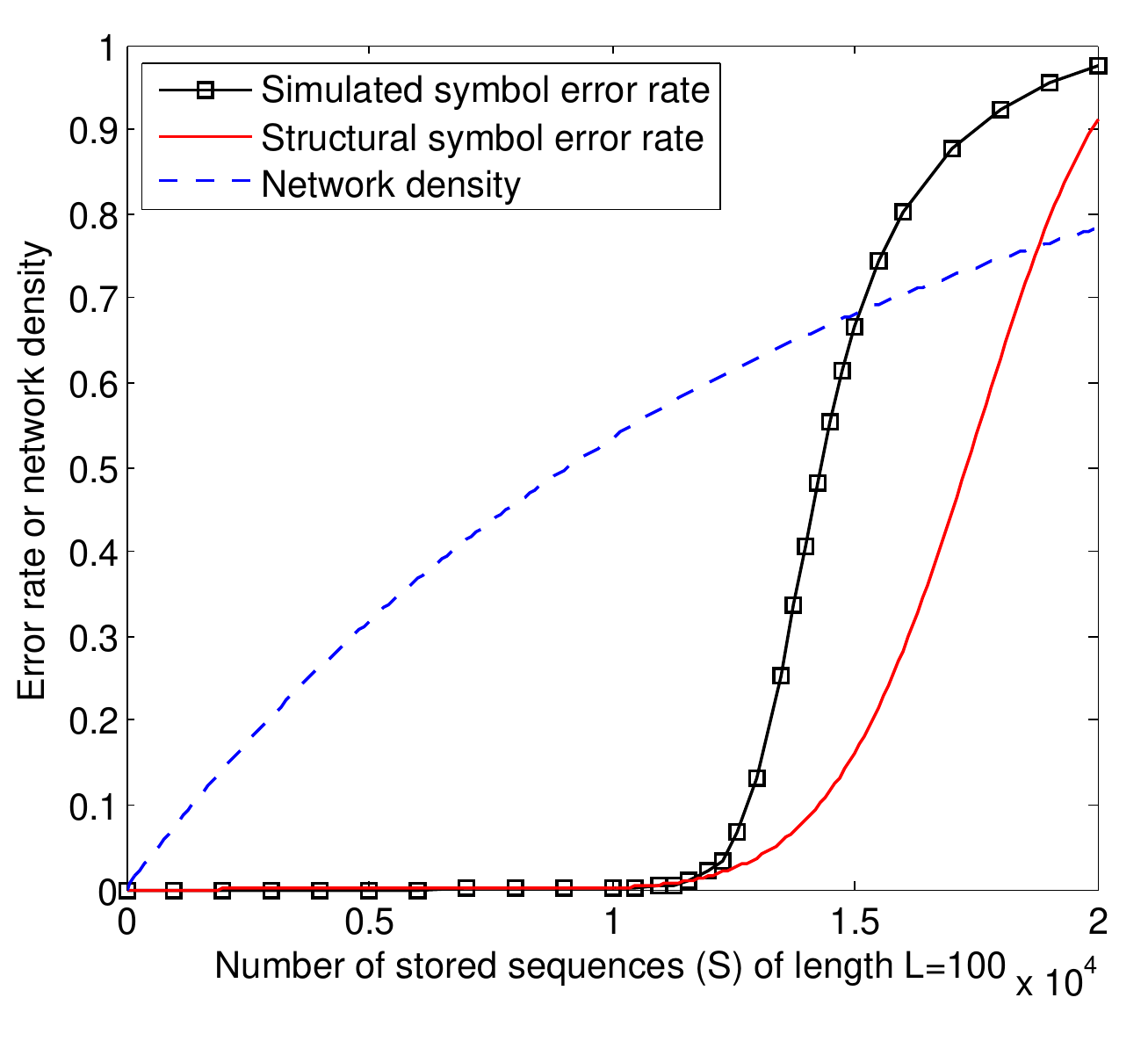}
\caption{Performance of the error tolerant decoding in a chain of tournaments with $\chi=20$, $l=256$ and $r=19$. The length of the sequences is $L=100$. Three curves are illustrated: simulated symbol error rate, structural symbol error rate and the network density.}
\label{error_rate_sequence_of_symbols}
\end{figure}

The theoretical error rate $P_{e, struct}$ has to be compared with the simulated SBER, knowing that we have always SBER $ \geq  P_{e, struct}$ . The result is shown in Fig. \ref{error_rate_sequence_of_symbols} for the tournament-based network with parameters $\chi=20$, $l=256$ and $r=19$ (that is with maximal temporal redundancy). The length of the sequences is $L=100$. We observe that the presence of errors in the previous positions do make the task more difficult for the decoder, but the performance is far from catastrophic. For example, for a structural SBER$=20\%$, the network is able to store 15000 sequences of length 100. By simulation, this same error rate corresponds to 13000 sequences.

\subsection{Sequence retrieval error rate}
As discussed before, giving a theoretical estimate of SBER is not trivial in sequential decoding, since the occurrence of one retrieval error at a given step will provide the next $r$ steps with erroneous information, and thus will potentially lead to an accumulated number of retrieval errors. We are thus interested in the sequence error rate (SQER), which is defined as the ratio of the number of incorrectly retrieved sequences to the total number of test sequences. Test sequences are among those having been stored by the network. The retrieval will be considered as a failure as soon as a single erroneous symbol appears.

Each decoding step is a small segmented problem: identify the target fanal by making use of the stimuli of the $r$ previous active fanals and the existing connections. As expressed in (\ref{innate_sym_err_rate}), the probability of making a right decision for a certain decoding step is $\left(1 - d^{r}\right)^{l - 1}$. There are $L-r$ steps for decoding a whole squence of length $L$. SQER is then estimated by the following formula:
\begin{equation}
   P_{e}^{\text{seq}} = 1 - \left(1 - d^{r}\right)^{(l - 1)(L - r)}.
\label{sqrer_seq_sym}
\end {equation}

We have now to answer the question: for a limited number of neural ressources ($n$ fanals), and a given learning set ($S$ sequences of length $L$), how can we optimize the number of clusters $\chi$ in order to minimize SQER? We fix $r=\chi -1$, since obviously it is of interest to take $r$ as large as possible to minimize the error rate, the density $d$ being independent of $r$ (see Section \ref{storage_seq_sym}).

Some assumptions may be made in order to simplify the demonstration. By supposing $d \ll 1$, we deduce from (\ref{den_seq_sym}) that
\begin{equation}
   d \approx  \frac{SL}{\chi l^2}.
\label{den_approx}
\end {equation}

For $r=\chi -1$, we have 
\begin{equation}
   \begin{aligned}
   P_{e}^{\text{seq}} &= 1 - {\left(1 - d^{\chi-1}\right)}^{(l - 1)(L - \chi +1)}\\
                                   & \approx l(L-\chi)d^{\chi} \text{, if $P_{e}^{\text{seq}} \ll 1$, $l \gg 1$ and $\chi \gg 1$}\\  
                                    &\approx l(L-\chi){\left(\frac{SL \chi }{n^2}\right)}^ {\chi }.                                   
   \end{aligned}
\label{sqrer_seq_sym_approx}
\end {equation}

By differentiating $\log(P_{e}^{\text{seq}})$ with respect to $\chi$, we obtain
\begin{equation}
 \begin{aligned}
       \frac{d\log(P_{e}^{\text{seq}})}{d\chi} & = \log\left( \frac{SL\chi}{n^2} \right)  - \frac{1}{\chi} +1 - \frac{1}{L-\chi}\\
 						             &\approx  \log\left( \frac{SL\chi}{n^2} \right)  +1 \text{, if $1 \ll \chi \ll L$.}\\  
 \end{aligned}
 \label{sqrer_seq_sym_deriv}
\end {equation}

Finally, setting the derivative equal to zero gives
\begin{equation}
      \chi_{opt} \approx \frac{n^2}{eSL}.
  \label{chi_opt}
\end {equation}

For example, for $n=4096, S=3000 \text{ and } L=100$, the optimal number of clusters is $\chi_{\text{opt}} \approx 20$, for a corresponding density $d \approx \frac{1}{e} \approx 0.37$. This is acceptable for a rough estimate, since with the same set of parameters, the exact formula of density (\ref{den_seq_sym}) gives $d \approx 0.30$. Note that $\chi_{\text{opt}}$ is proportional to the square of the number of fanals and inversely proportional to the sequence length and the number of sequences.

\subsection{Capacity and efficiency}

Each sequence is composed of $L \log_{2}(l)$ bits. If the number of stored sequences is small compared to the total number of possible ones, what we consider in the sequel, it has been shown in [\ref{GriRab12}] that the capacity, that is, the number of bits stored in the network approaches:
\begin{equation}
   C^{\text{seq}} = SL \log_{2}(l).
\label{capacity_seq_sym}
\end {equation}

Each cluster is connected via a vectorial arrow towards each of its $r$ downstream neighbors. A vectorial arrow is potentially composed of $l^2$ arrows, each of which can be encoded by 1 bit. Thus,  $Q$, the quantity of memory used by the network is:
\begin{equation}
   Q = r\chi l^{2}.
\label{memo_quant}
\end {equation}
With the same hypothesis as above, this leads to the expression of network efficiency:
\begin{equation}
 \begin{aligned}
   \eta^{\text{seq}} &= \frac{C}{Q} = \frac{SL \log_{2}(l)}{ r\chi l^{2}} \\
                                & \approx \frac{\log_{2}(n/\chi)}{e(\chi-1)}, \text{ for $r=\chi-1$.}
 \end{aligned}
\label{efficiency}
\end {equation}

For the same configuration as in the previous subsection, we have $\eta^{\text{seq}}=14.1\%$. This calculation is valid in the case of $r=\chi-1$, however, a higher efficiency is usually observed for a lower degree of anticipation $r$. Table \ref{table1} describes theoretical values for several different configurations of the network. For a reasonably sized network with the concern for biological plausibility, which means the number of fanals per cluster is of the order of 100, the efficiency is around $20\%$ to $30\%$. The efficiency tends extremely slowly to the asymptotical bound of $69\%$ [\ref{Knoblauch}] as the number of fanals tends to infinity.

\begin{table}[!t]
\centering
\caption{Maximum number of sequences (diversity) $S^{\text{seq}}$ that a chain of tournaments is able to store and retrieve with an error probability smaller than 0.01, for different values of $c$, $l$, $r$ and $L$. The values of corresponding efficiency $\eta^{\text{seq}}$ are also mentioned.}
\renewcommand{\arraystretch}{1.5}
%\begin{tabular}{cccccccc}
\begin{tabular}{p{0.6cm} p{0.6cm} p{0.6cm} p{0.6cm}| p{1.5cm} p{0.6cm}}
  \hline \hline
  $\chi$ & $l$ & $r$ & $L$ & $S^{\text{seq}}$ & $\eta^{\text{seq}}$ \\
  \hline \hline
  8 & 512 & 3 & 16 & 1513 & 3.5\%  \\
  50 & 128 & 10 & 100 & 2335 & 20.0\% \\
  50 & 128 & 20 & 100 & 5693 & 24.3\% \\
  50 & 128 & 49 & 100 & 11728 & 20.5\% \\
  30 & 512 & 23 & 100 & 57206 & 28.5\% \\
  30 & 512 & 29 & 100 & 70914 & 28.0\% \\ 
  100 & $2^{26}$ & 40 & 200 & $1.6.10^{15}$ & 45.1\% \\
  \hline
\end{tabular}
%\scalebox{0.5}{\usebox{\tablebox}}
\label{table1}
\end{table}

\section{Storage of vectorial sequences in generalized chain of tournaments}
\label{Chpt_learning_seq_pat}

The structure represented on the right in Fig. \ref{chain_of_cliques_vs_tournaments} can be viewed as a generalization of the clique-based networks. It becomes a clique by setting $ r = \chi - 1$ (two oriented connections being equivalent to a non oriented one). It is still possible to generalize furthermore this topology:
\begin{enumerate}
    \item A chain of tournaments is not necessarily a closed loop;
    \item A given component of the sequence at time $\tau$, $p_{\tau}$, is not necessarily a single symbol, but a set of parallel symbols that corresponds to a set of fanals in different clusters. The sequence then becomes vectorial.
\end{enumerate}

\subsection{Sparse random patterns}
\label{sparse_rand_pat}
\begin{figure}[!t]
\centering
\includegraphics[width=0.45\textwidth]{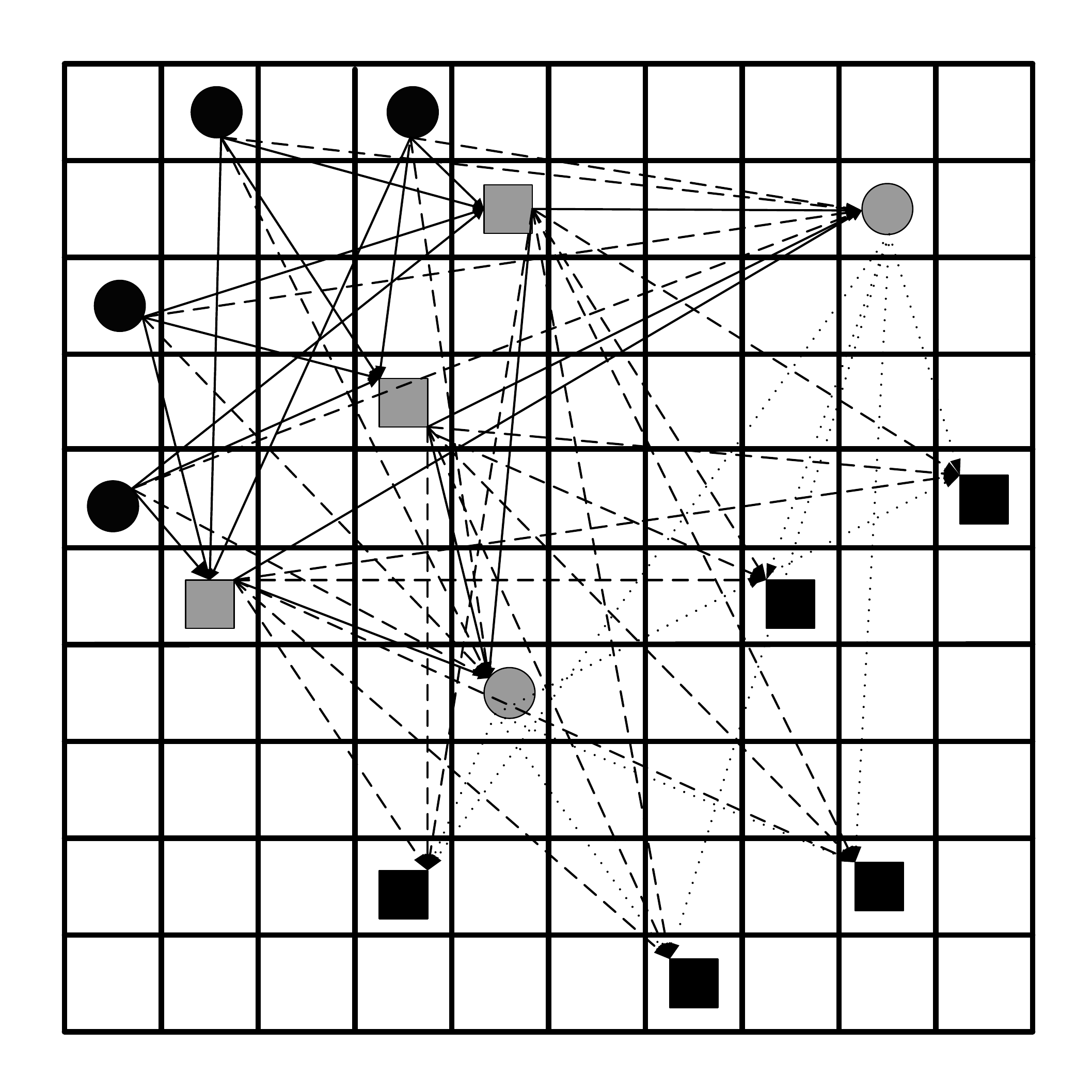}
\caption{Learning vectorial sequences in a network composed of 100 clusters by the generalized chain of tournaments. Clusters are represented by squares in the grid. Four patterns with different orders are represented: black circles, grey squares, grey circles and black squares. The incident degree is $r = 2$.}
\label{chain_of_tournaments_open}
\end{figure}

The authors of [\ref{Aliabadi}] have introduced a mechanism of storing sparse messages in networks of neural cliques in order to satisfy the ``sparse coding'' vision of mental information [\ref{Foldiak}] [\ref{Olshausen}]. A \textit{sparse message} or a \textit{sparse pattern} is a message that contains few expressed elements. Back to our sequential models, a sparse pattern is mapped to a set of $c$ fanals, with $c<\chi$, distributed in a certain number of clusters, with only one fanal per cluster. In Fig. \ref{chain_of_tournaments_open}, four sparse patterns with different orders are represented: black circles (order 4), grey squares (order 3), grey circles (order 2) and black squares (order 5). Here, we are not interested in the way to store and recall static sparse patterns, the different aspects of which are assessed in [\ref{Aliabadi}], but in the association between these patterns to build up sequential behavior.

The storage and the retrieval of sequences of sparse patterns can be implemented in the single layered structure illustrated in Fig. \ref{chain_of_tournaments_open}. Clusters are represented by squares in the grid. Each pattern generates activities in a limited number of distributed clusters. The elements of a single pattern are not associated by connections. On the other hand, two patterns that are linked are associated through an oriented complete bipartite graph. The succession of patterns is then carried by a chain of tournaments with degree $r$. In Fig. \ref{chain_of_tournaments_open}, we have $r = 2$, since the first pattern (filled circles) is connected (via solid arrows) to the second (grey squares) and to the third (grey circles, via dotted arrows), but not to the fourth (filled squares). This connectivity structure is similar to the sparse, distributed model of Rinkus [\ref{Rinkus1}], although the latter only considers connections between two consecutive patterns (no anticipation), and the way to organize clusters is different.

\subsection{Global decoding schemes}
\label{glb_decod_schm}

To retrieve the whole sequence, the network should be provided at least with a cue sequence of any $r$ successive patterns (a cue sequence of less than $r$ patterns makes the retrieval task not guaranteed but still possible), not necessarily at the beginning. The actual state of the network $\Phi_{\tau}$ is defined by the collection of the fanals that constitute $r$ estimated successive patterns at time $\tau$:
\begin{equation}
\label{network_state}
  \Phi_{\tau} = \bigcup_{t=\tau}^{\tau+r} {\tilde{p_{t}}} \text{ , with } \tilde{p_{t}}=\{ n_{ij}^{t}, n_{i'j'}^{t}, ...\}.
\end{equation}
All the fanals in $\Phi_{\tau}$ are activated:
\begin{equation}
\label{activation}
  \forall n_{ij} \in \Phi_{\tau} , v(n_{ij})=1 \text{ , otherwise } v(n_{ij})=0.
\end{equation}
A step of message passing is then done for all the fanals in the network, as defined by:
\begin{equation}
\label{sequence_message_passing}
   \forall i \in \llbracket 1,\chi \rrbracket, j \in \llbracket 1, l \rrbracket, v(n_{ij}) \leftarrow \displaystyle \sum_{i'=1}^{\chi} {w_{(i,j)(i',j')}v(n_{i' j'})}.
\end{equation}

Unlike the looped tournament-based networks in Section \ref{error_tolerant_decoding}, the emplacement of the target pattern is unknown a priori. As a consequence, one has to process a \textit{global} selection of winners rule instead of a \textit{local} one. Obviously, the local WTA automatically leaves one or several activated fanals in every cluster.

Several strategies are possible for a global selection. Three of them are presented here with the same example: after the message passing step, let us suppose that the set of 10 fanal scores is $\{v_{A}=5,v_{B}=6,v_{C}=1,v_{D}=8,v_{E}=7,v_{F}=7,v_{G}=8,v_{H}=5,v_{I}=0,v_{J}=8\}$ .
\begin{enumerate}
    \item Threshold Selection (TS): only the fanals with a score superior or equal to a predefined threshold $\theta$ are selected. For example, if $\theta=6$, the estimated pattern will be \{B,D,E,F,G,J\};
    \item Global Winner-Takes-All (GWTA): only the fanals with the maximal score are selected. The estimated pattern will be \{D,G,J\};
    \item Global Winners-Take-All (GWsTA): the $\alpha$ fanals with the maximal or near maximal scores are selected. If several fanals have the same score as the $\alpha^{\text{th}}$ does, these fanals are selected as well. For example, if $\alpha=4$, the estimated pattern will be \{D,E,F,G,J\}. In this case, the number of selected fanals can be slightly larger than the targeted number $\alpha$.
\end{enumerate}

The network state is then updated by replacing the oldest pattern in $\Phi$ with the most recently decoded pattern. The decoding process then repeats, starting from (\ref{network_state}).

TS is simple and efficient when the order $c$ of each pattern is identical. In this case, one can simply apply a threshold $\theta=rc$ to reach good retrieval performance. However, the performance of TS is very sensitive to the threshold chosen. In a general case, the optimal value of $\theta$ is $|\Phi|$, the cardinality of $\Phi$. This makes it not applicable to sequences with variable-order patterns, if one does not know beforehand the order of previous patterns. In a similar way, the choice of parameter $\alpha$ in GWsTA is dependent on the order of the target pattern, the optimal choice being $\alpha=c$. Nevertheless, compared to TS, GWsTA is much more flexible concerning variable-order patterns. Setting the value of $\alpha$ equal to the smallest order $c_{\text{min}}$ is generally good enough, if the order of patterns does not vary very much and $c_{\text{min}}$ is not too small.  On the other hand, GWTA is totally independent of the pattern order. The counterpart is that a fanal obtaining erroneously the highest score will extinguish all the rest.

\begin{figure}[!h]
\centering
\includegraphics[width=0.45\textwidth]{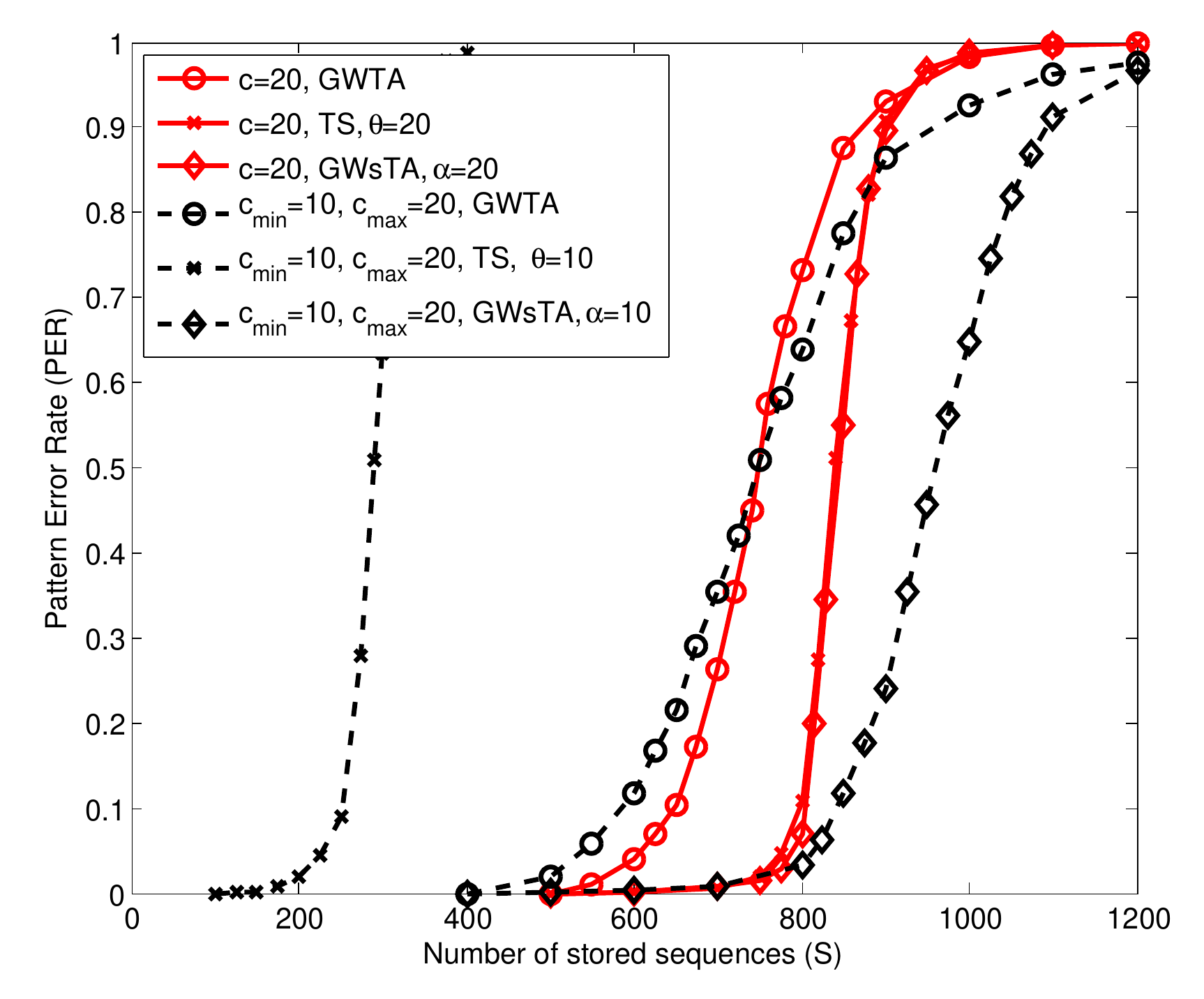}
\caption{Comparison of PER of vectorial sequences by applying different decoding schemes: TS, GWTA and GWsTA. The network is composed of 100 clusters of 64 fanals. The length of the sequences to store is 100. Two cases are simulated: all patterns contain 20 fanals; or the pattern order varies uniformly between 10 and 20. The parameter $\theta$ or $\alpha$ for the corresponding decoding scheme is properly chosen.}
\label{comp_decoding_algo_single_layer}
\end{figure}

Fig. \ref{comp_decoding_algo_single_layer} compares different decoding schemes in terms of Pattern Error Rate (PER): TS, GWTA and GWsTA. The simulated network is composed of 100 clusters of 64 fanals. The length of the sequences to store is 100. Two cases are simulated: all patterns contain 20 fanals; or the pattern order varies uniformly between 10 and 20. The parameters for each decoding scheme: $\theta$ or $\alpha$ are carefully chosen. When pattern order is fixed, TS and GWsTA have almost equivalent performance and both outperforms GWTA. With variable-order patterns, GWsTA with an optimally chosen $\alpha$ outperforms once again GWTA, while TS has very poor performance in this case. Overall, GWsTA is generally the best decoding scheme in terms of PER, while GWTA is more flexible, thus is a good tradeoff if the orders of patterns are completely unknown.

\subsection{Cluster activity restriction}
\label{cluster_act_restct}
The decoding strategies described in the previous section could be problematic if a fanal is shared by two or more patterns within range $r$.

As previously explained, the organization of the sequence storage involves two types of redundancy: spatial redundancy carried by the order of the pattern $c$ and temporal redundancy offered by anticipation degree $r$. The lack or weakness of one may be compensated by the other.  One may expect that sequences of small patterns can be better retrieved by increasing $r$, but in practice the suitable range of $r$ for a good performance is very limited. This issue is caused by the fact that $r$ successive patterns may share a certain number of fanals that lead to possible auto-connections. To take a simple example, consider the cue sequence  $p_{1} - p_{2} - ... - p_{r}$. The target pattern $p_{r+1}$ should be triggered due to the established connections between $p_{1} - p_{2} - ... - p_{r}$ and  $p_{r+1}$. Now suppose that $p_{1}$ and $p_{r}$ share exactly the same set of fanals. In this case, $p_{r}$ is also connected to the cue sequence $p_{1} - p_{2} - ... - p_{r}$, since by construction,  $p_{r}$ is already connected to $p_{1} - p_{2} - ... - p_{r-1}$ and $p_{1}=p_{r}$. As a result.  $p_{r+1}$ and  $p_{r}$ will both be triggered. A partial sharing of fanals among the $r$ successive patterns will cause the similar interference issue.  

To avoid this, one trick is to set a restriction on the cluster activity during the storage and consequently during the decoding as well: the clusters involved in a pattern are not allowed to participate in the activities of the $r$ following ones. It is then possible to estimate SQER. At each decoding step, we deduce the estimate at time $\tau+r+1$, denoted by $\tilde{p}_{\tau+r+1}$, from the $r$ previous estimates $\{\tilde{p_{t}}, t\in \llbracket \tau+1, \tau+r \rrbracket\}$, which are supposed to be correct: $\forall t\in \llbracket \tau+1, \tau+r \rrbracket, \tilde{p}_{t}=p_{t}$. Three groups of fanals should be distinguished for error probability estimation:
\begin{enumerate}
    \item all the fanals in the clusters involved in $\{p_{t}, t\in \llbracket \tau+1, \tau+r \rrbracket\}$, the number of which is $rl$. The probability of these fanals to be incorrectly stimulated is zero. In fact, cluster activity restriction forbids connections within the same cluster, that makes the scores of these fanals always less than $rc$.
    \item the fanals involved in the following patterns $\{p_{t}, t\in \llbracket \tau+r+2, \tau+2r+1\rrbracket \}$, the number of which is $(r-1)c$. These fanals deterministically receive stimuli from part of currently activated patterns. For example, the scores of the fanals in $p_{\tau+r+2}$ reach at least $(r-1)c$, since they are connected to $\{p_{t}, t\in \llbracket \tau+1, \tau+r \rrbracket \}$. As a consequence, $c$ spurious connections, the probability of each of them being $d$, are enough to make a wrong decision. The probability of these $(r-1)c$ fanals not to be incorrectly stimulated is thus $\prod_{i=1}^{r-1}{\left(1-d^{ic}\right)^{c}}$.
    \item all the remaining fanals except those in the target pattern $p_{\tau+r+1}$, the number of which is $n-rl-rc$. The probability of these fanals not to be incorrectly stimulated is $\left(1-d^{rc}\right)^{n-rl-rc}$.
\end{enumerate}

This estimation repeats $L-r$ times if the $r$ patterns at the beginning of the sequence are provided. Finally, SQER is estimated as:
\begin{equation}
\label{sqrer_cluster_act_restric}
   P_{e}=1-\left[\left(1-d^{rc}\right)^{n-rl-rc}\prod_{i=1}^{r-1}{\left(1-d^{ic}\right)^{c}}\right]^{L-r}
\end{equation}
with
\begin{equation}
\label{density_cluster_act_restric}
   d \approx 1-\left(1-\frac{rc^2}{n^2}\right)^{SL}.
\end{equation}

\begin{figure}
\centering
\includegraphics[width=0.5\textwidth]{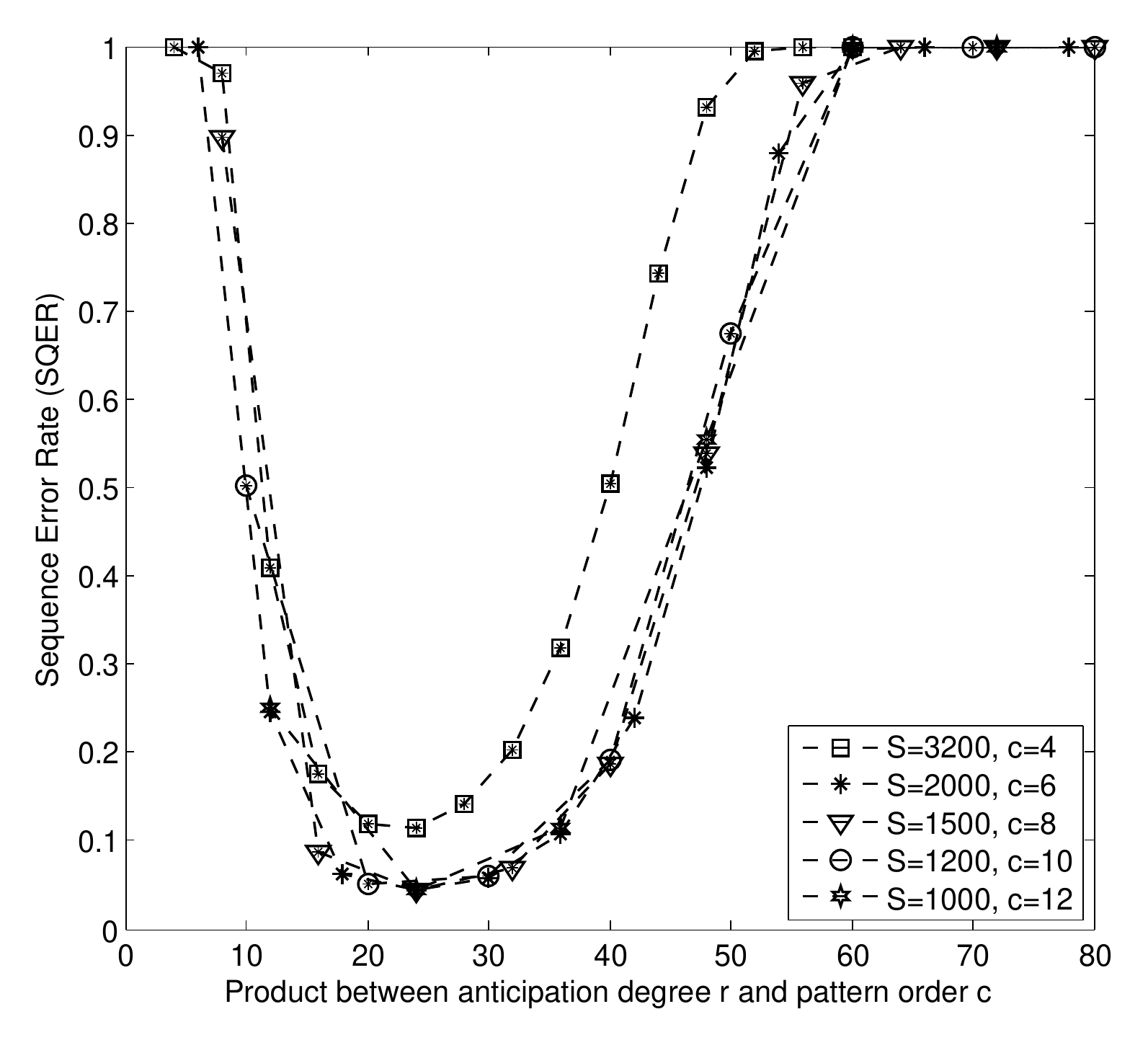}
\caption{Theoritical SQER in function of the the product between anticipation degree $r$ and pattern order $c$. The network is composed of 100 clusters of 64 fanals. The length of the sequences stored is 100. The curves correspond to sequences of patterns of different orders $c$: 4, 6, 8, 10 and 12. The number of sequences $S$ is adjusted for each $c$ in the way that the lowest point of each curve represents a SQER of about 1\%.}
\label{relation_optimal_r_c}
\end{figure}

Fig. \ref{relation_optimal_r_c} makes it clear that it is the product of $r$ and $c$ that determines the system retrieval performance. For example, in the network composed of 100 clusters of 64 fanals storing sequences of length 100, the optimal value of this product is approximately 25, and this is hardly dependent on different learning sets. Indeed, the product $rc$ corresponds to the quantity of available input information for the next pattern decoding. For sequences composed of patterns of large order, only a few previous instants of information are needed to decode the next pattern, whereas for patterns with small order more temporally distant information could be useful. However, an augmentation of either $c$ or $r$ will increase network density. A tradeoff should be made between the quantity of available input information and the density.

\section{Double layered structure}
\label{dbl_layer_strct}
\subsection{Error avalanche issue}
Up to now, the sequential decoding in tournament-based networks did not involve iterative processing. The overall sequence is encoded by connecting the group of fanals that represents a first pattern A to those that represent a second pattern B, which then connects to a third pattern C, etc. The problem of this simple chaining process is that an error in pattern A will potentially activate fanals that do not take part of pattern B, while some part of pattern B will not be stimulated. B is transformed into a corrupted version B'. Therefore, the retrieval of pattern C will be even more challenging and more subject to errors. This is called ``avalanche of errors''. We have discussed a similar issue in Section \ref{error_tolerant_decoding} with the decoding of sequences of symbols. Our solution was to use a sufficiently large value of $r$ in order to assure error tolerant decoding.

Some studies [\ref{kleinfeld}], [\ref{sompolinsky}] suggest theoretical solutions to the error avalanche problem, which are based on the error correction properties of auto-associative networks. Here, auto-association refers to the strengthening of synaptic connections between cells encoding the same memory. It was proposed that the error avalanche problem could be avoided if each hetero-associative step in the chaining process is followed by an auto-associative step, which potentially enables to transform a corrupted memory into the accurate version. But in these models, synaptic connections are weighted, which is totally different from our binary model.

\subsection{Structure}
\begin{figure}
\centering
\includegraphics[width=0.5\textwidth]{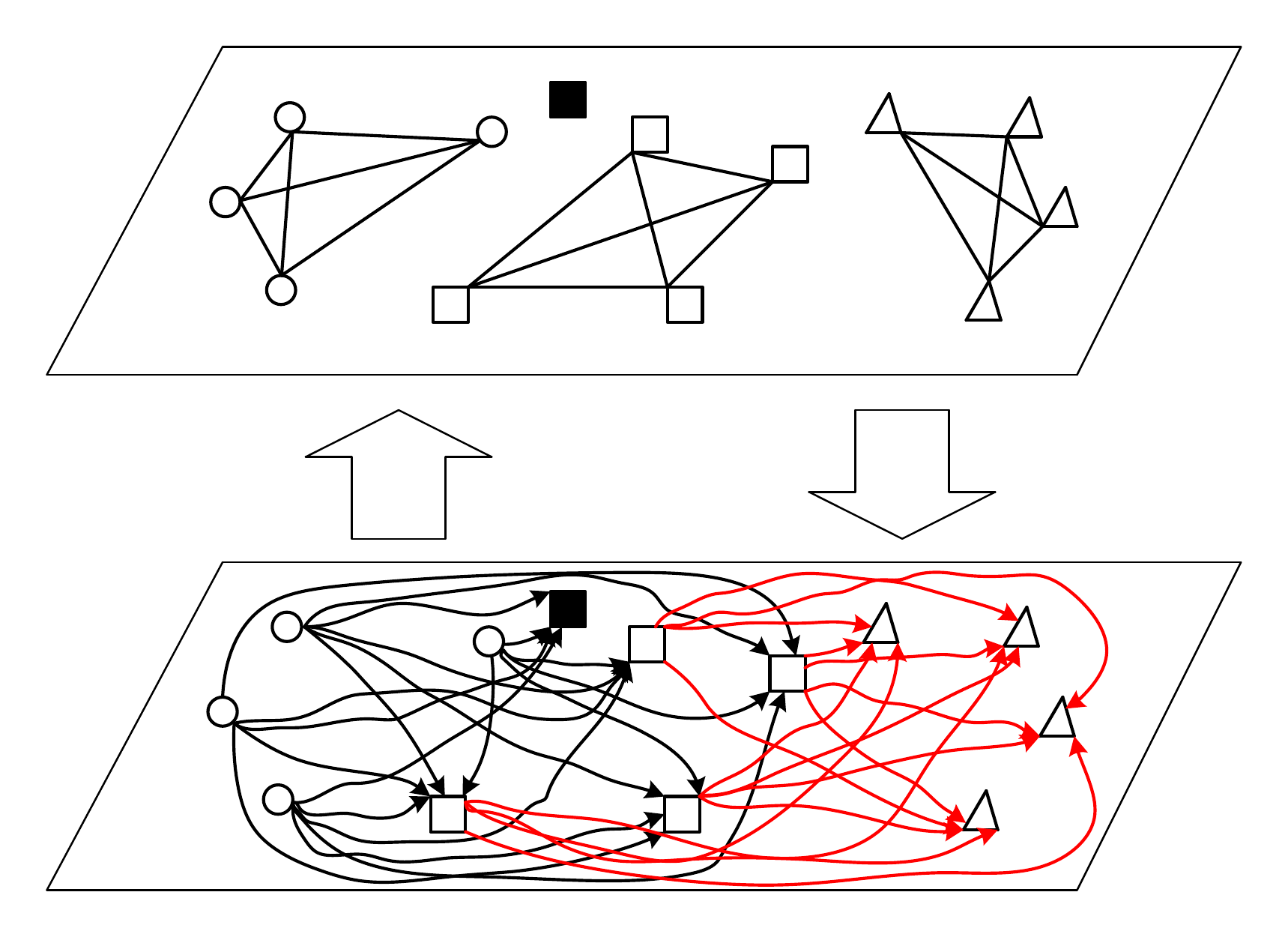}
\caption{Double layered structure. Upper: clique-based auto-associative layer for sparse pattern storing and decoding; lower: tournament-based hetero-associative layer for sequential storing and decoding. The interaction between these two layers assures accurate retrieval.}
\label{double_layered_structure}
\end{figure}

In a similar approach, we propose a double layered structure as illustrated in Fig. \ref{double_layered_structure}. The lower layer is \textit{tournament-based hetero-associative}, similar to the single layered network which stores sequential oriented associations between patterns. An upper layer of mirror fanals is superposed to emphasize the co-occurrence of the elements belonging to the same pattern, in the form of a clique. Therefore, this second layer is \textit{clique-based auto-associative}, similar to the sparse clique-based networks analyzed in [\ref{Aliabadi}]. 

\begin{figure}[!t]
\centering
\includegraphics[width=0.45\textwidth]{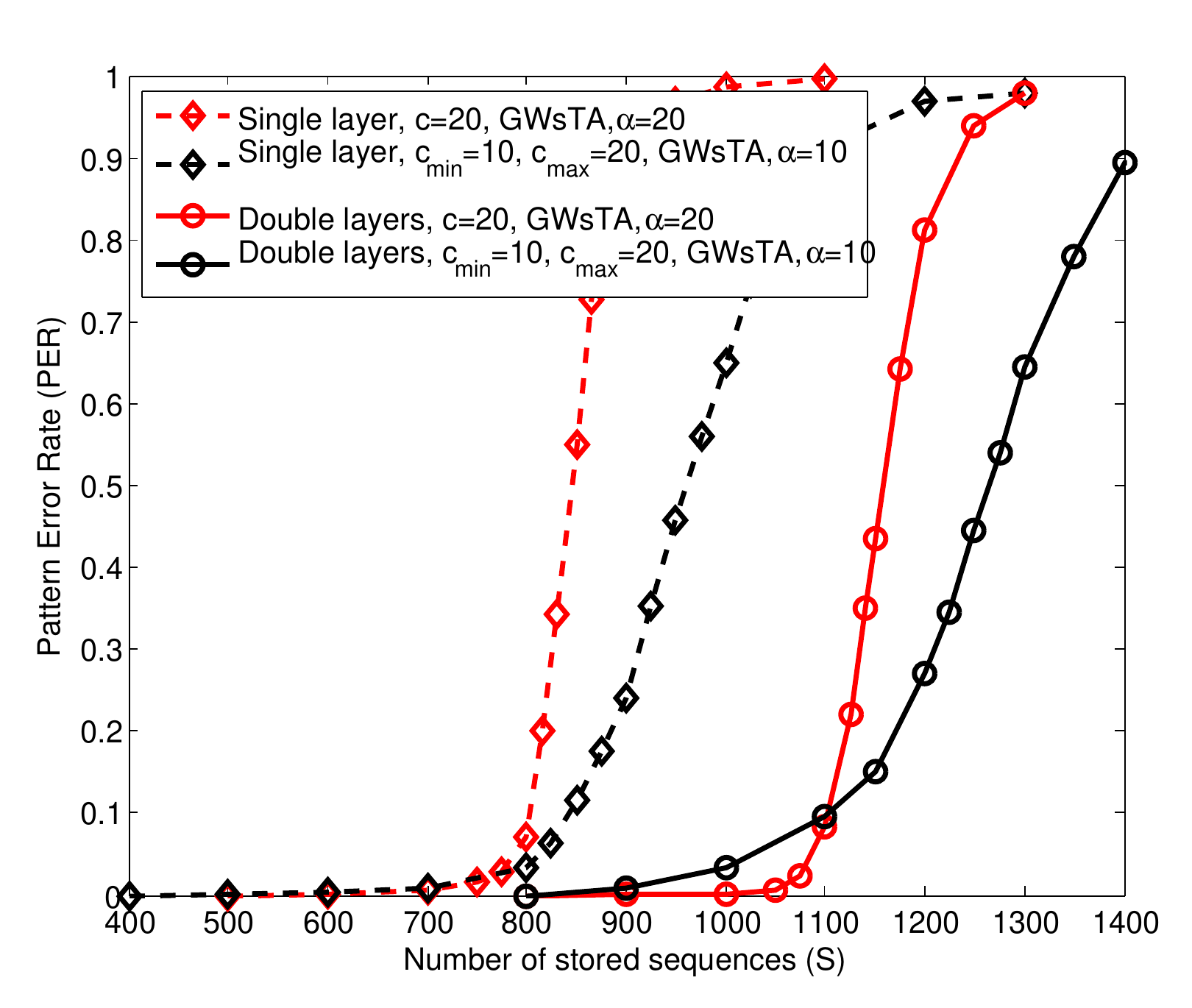}
\caption{Comparison of the single layered structure and the double layered structure in terms of PER. The network is composed of 100 clusters of 64 fanals. The length of the sequences stored is 100. Two cases are simulated: all patterns contain 20 fanals; or the pattern order varies uniformly from 10 to 20. Decoding scheme in clique-based layer: GWsTA with 4 iterations applied and memory effect $\gamma=1000$. Decoding scheme in tournament-based layer: GWsTA. $\alpha$ is optimally chosen for both layers. }
\label{comp_dbl_vs_single_layer}
\end{figure}

The clique-based auto-associative layer is supposed able to remove the insertions which are potentially activated because of the parasite connections in the tournament-based layer. For instance in Fig. \ref{double_layered_structure}, the upstream pattern represented by 4 circles activates the downstream pattern represented by 4 squares, but also an insertion, the filled square. The single layered network will in this case continue on decoding the following patterns, with the risk to amplify the error. On the contrary, the filled square does not belong to the clique in the clique-based layer, therefore will turn to be silent after some iterations. The accurate information is then retransmitted back to the tournament-based hetero-associative layer to pursue the sequential decoding. Here, an iterative GWsTA procedure is applied for clique decoding. Fig. \ref{comp_dbl_vs_single_layer} shows the gain in terms of PER of the double layered structure compared to the single layered structure.

However, one may note that the double layered structure uses more material than the single layered one. Efficiency is a more fair mesure for performance, which can be expressed as

\begin{equation}
\label{efficiency_seq_pat}
   \eta=\frac{SL(c\log_{2}l+\log_{2} \binom{\chi}{c})}{Q}.
\end{equation}

For the single layered structure, $Q_{\text{single}}=n^2$ since there are $n^2$ potential oriented connections each of which is encoded by 1 bit, whereas for the double layered one, $Q_{\text{double}} \approx 3 n^2 /2$, the number of non oriented connections being $\chi (\chi-1)l^2/2 \approx n^2/2$.  Even though, an efficiency gain is observed for the double layered structure. For instance, in the same configuration as Fig. \ref{comp_dbl_vs_single_layer}, with $c=20$, we have $\eta_{\text{double}}=32.2 \%$ compared to $\eta_{\text{single}}=26.7 \%$.

\subsection{Biological considerations}
The cooperation mechanism between the auto-associative and hetero-associative memory for accurate recalling is biologically plausible.  It has been suggested in  [\ref{Lisman}]  that in the hippocampus, the dentate and CA3 networks reciprocally cooperate for accurate recall of memory sequences. Auto-association occurs in CA3 to accentuate the simultaneity of elements in the same memory, whereas hetero-association occurs in the dentate to progress in the sequential decoding. Several anatomical and electrophysiological observations point out that dentate granule cells have axons called mossy fibers, which powerfully excite CA3 cells. And there is also a feedback information pathway from CA3 back to the dentate [\ref{Lisman2}].

However,  the realization of this cooperation by duplicating of identical nodes may appear irrelevant. Albeit mirror neurons as well as mirror systems have been directly observed in primate and other species, and it is speculated that this system provides the physiological mechanism for the perception/action coupling, it is not obvious at all that such a system also exists in the memory mechanism. To further approach biological reality, one may consider this structure as a single layer of unique nodes connected by relatively long and short synapses. The memory at one instant is represented by a pattern locally embedded in the terms of a neural clique. The information is exchanged rapidly between active nodes via short auto-associative synapses until the neural pattern stabilizes. Then this information is distributively spread via long hetero-associative synapses over the network to emerge the following local pattern, etc.

\section{Conclusion}
\label{conclu}

We have developed the possibility of tournament-based networks to process sequences of symbols or sparse patterns. Particularly, we have shown that they are able to store and retrieve a large number of long sequences, which could give them the ability to encode the flows with voluminous information, such as multimedia streams. As described in Section \ref{Chpt_learning_seq_pat}, the network made of 6400 nodes ($\chi=100$, $l=64$) is able to store about $S=700$ vectorial sequences composed of $L=100$ patterns of $c=20$ parallel symbols. Each pattern contains $c\log_2 l+\log_2 \binom{\chi}{c} \approx 188.9$ bits, which corresponds to a total information of about 13.2 Mbits. The double layered structure benefiting from iterative retrieving assures more accurate retrieving and higher efficiency. The decoding algorithm is error tolerant and anticipative, which mirrors what is known of biological neural networks.

Our model still possesses some degrees of freedom with respect to several network parameters, in particular the threshold $\theta$. Thresholds play a primordial role in the brain functioning. For instance, epilepsy is a common and diverse set of chronic neurological disorders, which result from abnormal, excessive neuronal activity in the brain. Some medications and treatments manage to increase the threshold in order to avoid this excessive activity. In our model, one may imagine implementing a non uniformly space-variant threshold laying over the whole network, which could be different from one cluster to another. The clusters with higher thresholds are thus penalized compared to those with lower ones. At the meantime, the threshold could also vary as long as the sequence progresses or be time-variant, and thus could ``guide" the sequence towards appropriate locations. This variable threshold should be a consequence of an intelligent training. 

Moreover, nothing by now enables to switch from one sequence to another to create some sort of ``association" or ``discrimination" between sequences. These associations, together with the choice of sequence paths mentioned above, are probably goal-oriented in the brain. Here once again, a variable threshold could be useful, since it could be encoded in the way such that this switching between sequences becomes possible. A speculation could be made that the level of intelligence would be, at least partially, based on how these kind of switchings or associations are performed in the brain. 

Finally, the principle depicted in Section \ref{dbl_layer_strct} could be extended to a hierarchical structure similar to that described in [\ref{Starzyk}] [\ref{Starzyk2}] for lecture principle, which is a four-level network that isolates letters, words, sentences, and strophes. Particularly, one can imagine a hierarchical network where each level passes a part of subsampled information to the next hierarchically higher level, which then aggregates received information in the form of super level cliques and/or chains of tournaments.

\end{document}